%% file: arxiv.tex
\documentclass[10pt,twocolumn,letterpaper]{article}

\usepackage{iccv}
\usepackage[format=plain,labelformat=simple,labelsep=period,font=small,compatibility=false]{caption}
\usepackage{times}
\usepackage{epsfig}
\usepackage{graphicx}
\usepackage{amsmath}
\usepackage{amssymb}
\usepackage{booktabs}
\usepackage{color}
\usepackage{xcolor}
\usepackage{algorithm}
\usepackage{algpseudocode}

\usepackage[pagebackref=true,breaklinks=true,letterpaper=true,colorlinks,bookmarks=false]{hyperref}

\usepackage[capitalize]{cleveref}
\crefname{section}{Sec.}{Secs.}
\Crefname{section}{Section}{Sections}
\Crefname{table}{Table}{Tables}
\crefname{table}{Tab.}{Tabs.}

\iccvfinalcopy %

\def\iccvPaperID{1917} %

\ificcvfinal\fi

\input{def}

\newcommand{\comment}[1]{\textcolor{gray}{#1}}

\begin{document}

\title{PhysDiff: Physics-Guided Human Motion Diffusion Model}

\author{
Ye Yuan \qquad Jiaming Song \qquad Umar Iqbal \qquad Arash Vahdat \qquad Jan Kautz \\[1mm]
NVIDIA \\
{ \url{https://nvlabs.github.io/PhysDiff}} \\
}

\maketitle
\ificcvfinal\thispagestyle{empty}\fi

\begin{abstract}
\vspace{-2mm}
Denoising diffusion models hold great promise for generating diverse and realistic human motions. However, existing motion diffusion models largely disregard the laws of physics in the diffusion process and often generate physically-implausible motions with pronounced artifacts such as floating, foot sliding, and ground penetration. This seriously impacts the quality of generated motions and limits their real-world application. To address this issue, we present a novel physics-guided motion diffusion model (PhysDiff), which incorporates physical constraints into the diffusion process. Specifically, we propose a physics-based motion projection module that uses motion imitation in a physics simulator to project the denoised motion of a diffusion step to a physically-plausible motion. The projected motion is further used in the next diffusion step to guide the denoising diffusion process. Intuitively, the use of physics in our model iteratively pulls the motion toward a physically-plausible space, which cannot be achieved by simple post-processing. Experiments on large-scale human motion datasets show that our approach achieves state-of-the-art motion quality and improves physical plausibility drastically ($>$78\% for all datasets).
\end{abstract}

\vspace{-3mm}
\section{Introduction}
\label{sec:intro}

Deep learning-based human motion generation is an important task with numerous applications in animation, gaming, and virtual reality. In common settings such as text-to-motion synthesis, we need to learn a conditional generative model that can capture the multi-modal distribution of human motions. The distribution can be highly complex due to the high variety of human motions and the intricate interaction between human body parts.
Denoising diffusion models~\cite{sohl-dickstein2015deep,ho2020denoising,song2020denoising} are a class of generative models that are especially suited for this task due to their strong ability to model complex distributions, which has been demonstrated extensively in the image generation domain~\cite{saharia2022photorealistic,ramesh2022hierarchical,rombach2022high, dhariwal2021diffusion}.
These models have exhibited strong mode coverage often indicated by high test likelihood~\cite{song2020score, kingma2021variational, vahdat2021score}.
They also have better training stability compared to generative adversarial networks (GANs~\cite{goodfellow2014generative}) and better sample quality compared to variational autoencoders (VAEs~\cite{kingma2013auto}) and normalizing flows~\cite{sinha2021d2c,vahdat2021score,vahdat2020nvae,aneja2021contrastive}. Motivated by this, recent works have proposed motion diffusion models~\cite{tevet2022human,zhang2022motiondiffuse} which significantly outperform standard deep generative models in motion generation performance.

However, existing motion diffusion models overlook one essential aspect of human motion -- the underlying laws of physics. Even though diffusion models have a superior ability to model the distribution of human motion, they still have no explicit mechanisms to enforce physical constraints or model the complex dynamics induced by forces and contact. As a result, the motions they generate often contain pronounced artifacts such as floating, foot sliding, and ground penetration. This severely hinders many real-world applications such as animation and virtual reality, where humans are highly sensitive to the slightest clue of physical inaccuracy~\cite{reitsma2003perceptual,hoyet2012push}. In light of this, a critical problem we need to address is making human motion diffusion models physics-aware.

\begin{figure}
    \centering
    \includegraphics[width=\linewidth]{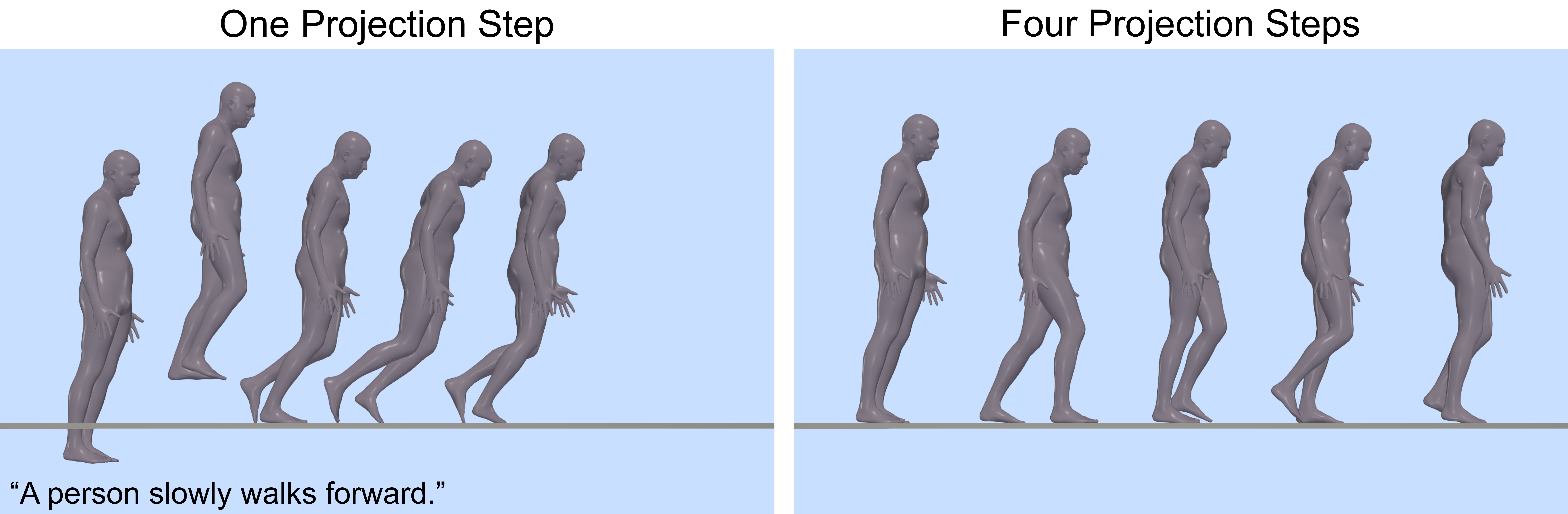}
    \vspace{-6mm}
    \caption{(\textbf{Left}) Performing one physics-based projection step (post-processing) at the end yields unnatural motion since the motion is too physically-implausible to correct. (\textbf{Right}) Our approach solves this issue by iteratively applying physics and diffusion.}
    \label{fig:intro}
    \vspace{-4mm}
\end{figure}

To tackle this problem, we propose a novel physics-guided motion diffusion model (PhysDiff) that instills the laws of physics into the denoising diffusion process. Specifically, PhysDiff leverages a physics-based motion projection module (details provided later) that projects an input motion to a physically-plausible space. During the diffusion process, we use the motion projection module to project the denoised motion of a diffusion step into a physically-plausible motion. This new motion is further used in the next diffusion step to guide the denoising diffusion process. Note that it may be tempting to only add the physics-based projection at the end of the diffusion process, i.e., using physics as a post-processing step. However, this can produce unnatural motions since the final denoised kinematic motion from diffusion may be too physically-implausible to be corrected by physics (see Fig.~\ref{fig:intro} for an example) and the motion may be pushed away from the data distribution. Instead, we need to embed the projection in the diffusion process and apply physics and diffusion iteratively to keep the motion close to the data distribution while moving toward the physically-plausible space (see Sec.~\ref{sec:vs_postproc}).

The physics-based motion projection module serves the vital role of enforcing physical constraints in PhysDiff, which is achieved by motion imitation in a physics simulator. Specifically, using large-scale motion capture data, we train a motion imitation policy that can control a character agent in the simulator to mimic a vast range of input motions. The resulting simulated motion enforces physical constraints and removes artifacts such as floating, foot sliding, and ground penetration. Once trained, the motion imitation policy can be used to mimic the denoised motion of a diffusion step to output a physically-plausible motion.

We evaluate our model, PhysDiff, on two tasks: text-to-motion generation and action-to-motion generation. Since our approach is agnostic to the specific instantiation of the denoising network used for diffusion, we test two state-of-the-art (SOTA) motion diffusion models (MDM \cite{tevet2022human} and MotionDiffuse~\cite{zhang2022motiondiffuse}) as our model's denoiser. For text-to-motion generation, our model outperforms SOTA motion diffusion models significantly on the large-scale HumanML3D~\cite{guo2022generating} benchmark, reducing physical errors by more than 86\% while also improving the motion quality by more than 20\% as measured by the Frechet inception distance (FID). For action-to-motion generation, our model again improves the physics error metric by more than 78\% on HumanAct12~\cite{guo2020action2motion} and 94\% on UESTC~\cite{ji2018large} while also achieving competitive FID scores.

We further perform extensive experiments to investigate various schedules of the physics-based projection, \ie, at which diffusion timesteps to perform the projection. Interestingly, we observe a trade-off between physical plausibility and motion quality when varying the number of physics-based projection steps. Specifically, while more projection steps always lead to better physical plausibility, the motion quality increases before a certain number of steps and decreases after that, \ie, the resulting motion satisfies the physical constraints but still may look unnatural. This observation guides us to use a balanced number of physics-based projection steps where both high physical plausibility and motion quality is achieved. We also find that adding the physics-based projection to late diffusion steps performs better than early steps. We hypothesize that motions from early diffusion steps may tend toward the mean motion of the training data and the physics-based projection could push the motion further away from the data distribution, thus hampering the diffusion process. Finally, we also show that our approach outperforms physics-based post-processing (single or multiple steps) in motion quality and physical plausibility significantly.

Our contributions are summarized as follows:
\vspace{-1.5mm}
\begin{itemize}
\itemsep-1mm 
    \item We present a novel physics-guided motion diffusion model that generates physically-plausible motions by instilling the laws of physics into the diffusion process. Its plug-and-play nature makes it flexible to use with different kinematic diffusion models.
    \item We propose to leverage human motion imitation in a physics simulator as a motion projection module to enforce physical constraints.
    \item Our model achieves SOTA performance in motion quality and drastically improves physical plausibility on large-scale motion datasets. Our extensive analysis also provides insights such as schedules and tradeoffs, and we demonstrate significant improvements over physics-based post-processing.
\end{itemize}

\section{Related Work}
\label{sec:related_work}

\noindent\textbf{Denoising Diffusion Models.} Score-based denoising diffusion models~\cite{sohl-dickstein2015deep,ho2020denoising,song2020score,song2020denoising} have achieved great successes in various applications such as image generation~\cite{saharia2022photorealistic,ramesh2022hierarchical,rombach2022high,sinha2021d2c,vahdat2021score}, text-to-speech synthesis~\cite{kong2020diffwave}, 3D shape generation~\cite{zhou2021shape,luo2021diffusion,zeng2022lion}, machine learning security~\cite{nie2022DiffPure}, as well as human motion generation~\cite{tevet2022human,zhang2022motiondiffuse,ren2022diffusion}. These models are trained via denoising autoencoder objectives that can be interpreted as score matching~\cite{Vincent11}, and generate samples via an iterative denoising procedure that may use stochastic updates~\cite{bao2022analytic,bao2022estimating,dockhorn2021score,zhang2022gddim} which solve stochastic differential equations (SDEs) or deterministic updates~\cite{song2020denoising,lu2022dpm,liu2022pseudo,zhang2022fast,karras2022elucidating,dockhorn2022genie} which solve ordinary differential equations (ODEs).  

To perform conditional generation, the most common technique is classifier(-free) guidance~\cite{dhariwal2021diffusion,ho2022classifier}. However, it requires training the model specifically over paired data and conditions. Alternatively, one could use pretrained diffusion models that are trained only for unconditional generation. For example, SDEdit~\cite{meng2021sdedit} modifies the initialization of the diffusion model to synthesize or edit an existing image via colored strokes. In image domains, various methods solve linear inverse problems by repeatedly injecting known information to the diffusion process~\cite{lugmayr2022repaint,chung2021come,song2020score,choi2021ilvr,kawar2021snips,kawar2022denoising}. A similar idea is applied to human motion diffusion models in the context of motion infilling~\cite{tevet2022human}. 
In our case, generating physically-plausible motions with diffusion models has a different set of challenges. First, the constraint is specified through a physics simulator, which is non-differentiable. Second, the physics-based projection itself is relatively expensive to compute, unlike image-based constraints which use much less compute than the diffusion model in general. As a result, we cannot simply apply the physics-based projection to every step of the sampling process.

\vspace{2mm}
\noindent\textbf{Human Motion Generation.}
Early work on motion generation adopts deterministic human motion modeling which only generates a single motion~\cite{fragkiadaki2015recurrent,li2017auto,ghosh2017learning,martinez2017human,pavllo2018quaternet,gopalakrishnan2019neural,aksan2019structured,wang2019imitation,mao2019learning,yuan2019ego}. Since human motions are stochastic in nature, more work has started to use deep generative models which avoid the mode averaging problem common in deterministic methods. These methods often use GANs or VAEs to generate motions from various conditions such as past motions~\cite{barsoum2018hp,yan2018mt,aliakbarian2020stochastic,yuan2020residual,yuan2020dlow}, key frames~\cite{he2022nemf}, music~\cite{li2021ai,zhuang2022music2dance,li2022danceformer}, text~\cite{ahuja2019language2pose,bhattacharya2021text2gestures,guo2022generating,petrovich22temos}, and action labels~\cite{guo2020action2motion,petrovich2021action,cervantes2022implicit}. Recently, denoising diffusion models~\cite{sohl-dickstein2015deep,ho2020denoising,song2020denoising} have emerged as a new class of generative models that combine the advantages of standard generative models. Therefore, several motion diffusion models~\cite{tevet2022human,zhang2022motiondiffuse,ren2022diffusion} have been proposed which demonstrate SOTA motion generation performance. However, existing motion diffusion models often produce physically-implausible motions since they disregard physical constraints in the diffusion process. Our method addresses this problem by guiding the diffusion process with a physics-based motion projection module.

\vspace{2mm}
\noindent\textbf{Physics-Based Human Motion Modeling.}
Physics-based human motion imitation is first applied to learning locomotion skills such as walking, running, and acrobatics with deep reinforcement learning (RL)~\cite{liu2017learning,liu2018learning,merel2017learning,peng2018deepmimic,peng2018sfv}. RL-based motion imitation has also been used to learn user-controllable policies for character animation~\cite{bergamin2019drecon,park2019learning,won2020scalable}. For 3D human pose estimation, recent work has adopted physics-based trajectory optimization~\cite{zell2017joint,rempe2020,shimada2020physcap,shimada2021neural} and motion imitation~\cite{yuan20183d,yuan2019ego,isogawa2020optical,yuan2021simpoe,yi2022physical,luo2021dynamics,luo2022embodied} to model human dynamics. Unlike previous work, we explore the synergy between physics simulation and diffusion models, and show that applying physics and diffusion iteratively can generate more realistic and physically-plausible motions.

\section{Method}
\label{sec:method}

\begin{figure*}
    \centering
    \includegraphics[width=\linewidth]{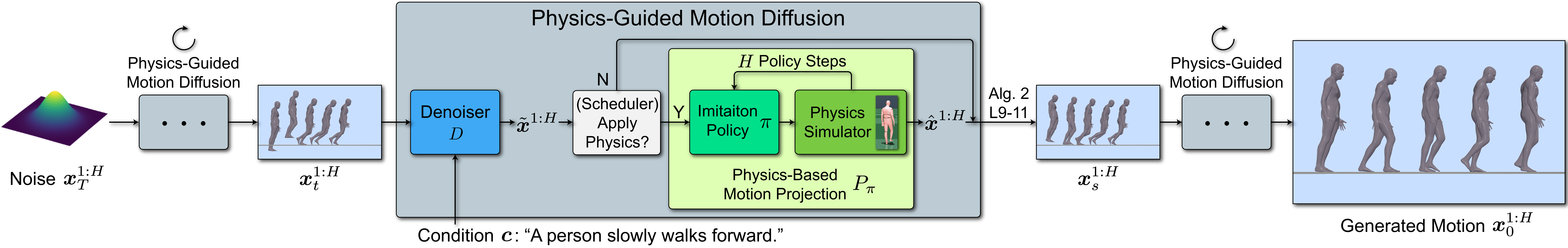}
    \caption{\textbf{Overview of PhysDiff.} Each physics-guided diffusion step denoises a motion from timestep $t$ to $s$, where physics-based motion projection is used to enforce physical constraints. The projection is achieved using a motion imitation policy to control a character in a physics simulator. A scheduler controls when the physics-based projection is applied. The denoiser can be any motion-denoising network.}
    \label{fig:overview}
    \vspace{-4mm}
\end{figure*}

Given some conditional information $\bs{c}$ such as text or an action label, we aim to generate a physically-plausible human motion $\bs{x}^{1:H} = \{\bs{x}^h\}_{h=1}^H$ of length $H$. Each pose $\bs{x}^h \in \mathbb{R}^{J \times D}$ in the generated motion is represented by the $D$-dimensional features of $J$ joints, which can be either the joint positions or angles. We propose a physics-guided denoising diffusion model (PhysDiff) for human motion generation. Starting from a noisy motion $\bs{x}_T^{1:H}$, PhysDiff models the denoising distribution $q(\bs{x}_{s}^{1:H}|\bs{x}_t^{1:H}, \mathcal{P}_\pi,\bs{c})$ that denoises the motion from diffusion timestep $t$ to $s$ ($s<t$). Iteratively applying the model denoises the motion into a clean motion $\bs{x}_0^{1:H}$, which becomes the final output $\bs{x}^{1:H}$. A critical component in the model is a physics-based motion projection module $\mathcal{P}_\pi$ that enforces physical constraints. It leverages a motion imitation policy $\pi$ to mimic the denoised motion of a diffusion step in a physics simulator and uses the simulated motion to further guide the diffusion process. An overview of our PhysDiff model is provided in Fig.~\ref{fig:overview}. In the following, we first introduce the physics-guided motion diffusion process in Sec.~\ref{method:sec:diffusion}. We then describe the details of the physics-based motion projection $\mathcal{P}_\pi$ in Sec.~\ref{method:sec:physics}.

\subsection{Physics-Guided Motion Diffusion}
\label{method:sec:diffusion}

\noindent\textbf{Motion Diffusion.} 
To simplify notations, here we sometimes omit the explicit dependence over the condition $\bs{c}$. Note that we can always train diffusion models with some condition $\bs{c}$; even for the unconditional case, we can condition the model on a universal null token $\varnothing$~\cite{ho2022classifier}.

Let $p_0(\bs{x})$ denote the data distribution, and define a series of time-dependent distributions $p_t(\bs{x}_t)$ by injecting \textit{i.i.d.} Gaussian noise to samples from $p_0$, \ie, $p_t(\bs{x}_t | \bs{x}) = \mathcal{N}(\bs{x}, \sigma_t^2 \mathbf{I})$, where $\sigma_t$ defines a series of \textit{noise levels} that is increasing over time such that $\sigma_0 = 0$ and $\sigma_T$ for the largest possible $T$ is much bigger than the data's standard deviation. Generally, diffusion models draw samples by solving the following stochastic differential equation (SDE) from $t=T$ to $t=0$ \cite{grenander1994representations,karras2022elucidating,zhang2022gddim}:
\begin{align}
    \mathrm{d} \bs{x} = - (\beta_t + \dot{\sigma}_t) \sigma_t \nabla_{\bs{x}} \log p_t(\bs{x}) \mathrm{d} t + \sqrt{2 \beta_t} \sigma_t \mathrm{d} \omega_t, \label{eq:score-sde}
\end{align}
where $\nabla_{\bs{x}} \log p_t(\bs{x})$ is the score function, $\omega_t$ is the standard Wiener process, and $\beta_t$ controls the amount of stochastic noise injected in the process; when it is zero, the SDE becomes and ordinary differential equation (ODE). A notable property of the score function $\nabla_{\bs{x}_t} \log p_t(\bs{x}_t)$ is that it recovers the minimum mean squared error (MMSE) estimator of $\bs{x}$ given $\bs{x}_t$ \cite{stein1981estimation,efron2011tweedie,saremi2019neural}:
\begin{align}
    \kin{\bs{x}} := \mathbb{E}[\bs{x} | \bs{x}_t] = \bs{x}_t + \sigma_t^2 \nabla_{\bs{x}_t} \log p_t(\bs{x}_t), \label{eq:tweedie} %
\end{align}
where we can essentially treat $\kin{\bs{x}}$ as a ``denoised'' version of $\bs{x}_t$. Since $\bs{x}_t$ and $\sigma_t$ are known during sampling, we can obtain $\nabla_{\bs{x}_t} \log p_t(\bs{x}_t)$ from $\kin{\bs{x}}$, and vice versa.

Diffusion models approximate the score function with the following denoising autoencoder objective~\cite{Vincent11}:
\begin{align}
    \mathbb{E}_{\bs{x} \sim p_0(\bs{x}), t \sim p(t), \epsilon \sim p(\epsilon)}[\lambda(t) \Vert \bs{x} - D(\bs{x} + \sigma_t \epsilon, t, \bs{c}) \Vert_2^2]
\end{align}
    where $D$ is the denoiser that depends on the noisy data, the time $t$ and the condition $\bs{c}$, $\epsilon \sim \mathcal{N}(\bs{0}, \bs{I})$, $p(t)$ is a distribution from which time is sampled, and $\lambda(t)$ is the loss weighting factor. The optimal solution to $D$ would be one that recovers the MMSE estimator $\kin{\bs{x}}$ according to \cref{eq:tweedie}. For a detailed characterization of the training procedure, we refer the reader to Karras \etal \cite{karras2022elucidating}.

After the denoising diffusion model has been trained, one can apply it to solve the SDE / ODE in \cref{eq:score-sde}. A particular approach, DDIM~\cite{song2020denoising}, performs a one-step update from time $t$ to time $s$ ($s < t$) given the sample $\bs{x}_t$, which is described in~\cref{algo:ddim}. Intuitively, the sample $\bs{x}_s$ at time $s$ is generated from a Gaussian distribution; its mean is a linear interpolation between $\bs{x}_t$ and the denoised result $\kin{\bs{x}}$, and its variance depends on a hyperparameter $\eta \in [0, 1]$.
Specifically, $\eta = 0$ corresponds to denoising diffusion probabilistic models (DDPM~\cite{ho2020denoising}) and $\eta = 1$ corresponds to denoising diffusion implicit models (DDIM~\cite{song2020denoising}). We find $\eta = 0$ produces better performance for our model.
Since the above sampling procedure is general, it can also be applied to human motion data, \ie, $\bs{x}^{1:H}$. To incorporate the condition $\bs{c}$ during sampling, one can employ classifier-based or classifier-free guidance~\cite{dhariwal2021diffusion,ho2022classifier}.

\begin{algorithm}
    \caption{DDIM sampling algorithm}
    \label{algo:ddim}
    \begin{algorithmic}[1]
    \State \textbf{Input}: Denoiser $D$, sample $\bs{x}_t$ at time $t$, target time $s$, condition $\bs{c}$, hyperparameter $\eta \in [0, 1]$.
    \State Compute the denoised result $\kin{\bs{x}} := D(\bs{x}_t, t, \bs{c})$.
    \State Obtain variance $v_s$ as a scalar that depends on $\eta$.
    \State Obtain mean $\mu_s$ as a linear combination of $\kin{\bs{x}}$ and $\bs{x}_t$:
    $$
        \mu_s := \kin{\bs{x}} + \frac{\sqrt{\sigma_s^2 - v_s}}{\sigma_t} (\bs{x}_t - \kin{\bs{x}})
    $$
    \State Draw sample $\bs{x}_s \sim \mathcal{N}(\mu_s, v_s \bs{I})$.
    \end{algorithmic}
\end{algorithm}

\noindent\textbf{Applying Physical Constraints.}
Existing diffusion models for human motions are not necessarily trained on data that complies with physical constraints, and even if they are, there is no guarantee that the produced motion samples are still physically realizable, due to the approximation errors in the denoiser networks and the stochastic nature of the sampling process. While one may attempt to directly correct the final motion sample to be physically-plausible, the physical errors in the motion might be so large that even after such a correction, the motion is still not ideal (see Fig.~\ref{fig:intro} for a concrete example and Sec.~\ref{sec:vs_postproc} for comparison).

To address this issue, we exploit the fact that diffusion models produce intermediate estimates of the desired outcome, \ie, the denoised motion $\kin{\bs{x}}^{1:H}$ of each diffusion step. In particular, we may apply physical constraints not only to the final step of the diffusion process, but to the intermediate steps as well. Concretely, we propose a physics-based motion projection $\mathcal{P}_\pi: \mathbb{R}^{H \times J \times D} \to \mathbb{R}^{H \times J \times D}$ as a module that maps the original motion $\kin{\bs{x}}^{1:H}$ to a physically-plausible one, denoted as $\phys{\bs{x}}^{1:H}$. We incorporate the proposed physics-based projection $\mathcal{P}_\pi$ into the denoising diffusion sampling procedure, where the one-step update from time $t$ to time $s$ is described in \cref{algo:physdiff}. The process differs from the DDIM sampler mostly in terms of performing the additional physics-based projection; as we will show in the experiments, this is a simple yet effective approach to enforcing physical constraints. Notably, our model, \mbox{PhysDiff}, is agnostic to the specific instantiation of the denoiser $D$, which is often implemented with various network architectures. The physics-based projection in PhysDiff is also only applied during diffusion sampling (inference), which makes PhysDiff generally compatible with different pretrained motion diffusion models. In other words, PhysDiff can be used to improve the physical plausibility of existing diffusion models without retraining.

\vspace{-1mm}
\begin{algorithm}[H]
    \caption{PhysDiff sampling algorithm for motion.}
    \label{algo:physdiff}
    \begin{algorithmic}[1]
    \State \textbf{Input}: Denoiser $D$, sample $\bs{x}_t^{1:H}$ at time $t$, condition $\bs{c}$, target time $s$, physics-based projection $\mathcal{P}_\pi$, $\eta \in [0, 1]$.
    \State Compute the denoised motion $\kin{\bs{x}}^{1:H} := D(\bs{x}^{1:H}_t, t, \bs{c})$.
    \If{projection is performed at time $t$}
    \State $\phys{\bs{x}}^{1:H} := \mathcal{P}_\pi(\kin{\bs{x}}^{1:H})$ \comment{\quad\# Physics-Based Projection}
    \Else  
    \State $\phys{\bs{x}}^{1:H} := \kin{\bs{x}}^{1:H}$
    \EndIf
    \State \comment{\# The remaining part is similar to DDIM}
    \State Obtain variance $v_s$ as a scalar that depends on $\eta$.
    \State Obtain mean $\mu_s$:
    $$
        \mu_s := \phys{\bs{x}}^{1:H} + \frac{\sqrt{\sigma_s^2 - v_s}}{\sigma_t} (\bs{x}_t^{1:H} - \phys{\bs{x}}^{1:H})
    $$
    \State Draw sample $\bs{x}^{1:H}_s \sim \mathcal{N}(\mu_s, v_s \bs{I})$.
    \end{algorithmic}
\end{algorithm}

\vspace{-2mm}
\noindent\textbf{Scheduling Physics-based Projection.}
Due to the use of physics simulation, the projection $\mathcal{P}_\pi$ is rather expensive, and it is infeasible to perform the projection at every diffusion timestep. Therefore, if we have a limited number of physics-based projection steps to be performed, we need to prioritize certain timesteps over others. Here, we argue that we should not perform physics projection when the diffusion noise level is high. This is because the denoiser $D$ by design gives us $\kin{\bs{x}}^{1:H} = \mathbb{E}[\bs{x}^{1:H} | \bs{x}^{1:H}_t]$, \ie, the mean motion of $\bs{x}^{1:H}$ given the current noisy motion $\bs{x}^{1:H}_t$, and it is close to the mean of the training data for high noise levels when the condition $\bs{x}^{1:H}_t$ contains little information. Empirically, the mean motion often has little body movement due to mode averaging while still having some root translations and rotations, which is clearly physically-implausible. Correcting such a physically-incorrect motion with the physics-based projection would push the motion further away from the data distribution and hinder the diffusion process.
In Sec.~\ref{sec:schedule}, we perform a systematic study that validates this hypothesis and reveals a favorable scheduling strategy that balances sample quality and efficiency.

\subsection{Physics-Based Motion Projection}
\label{method:sec:physics}
An essential component in the physics-guided diffusion process is the physics-based motion projection $\mathcal{P}_\pi$. It is tasked with projecting the denoised  motion $\kin{\bs{x}}^{1:H}$ of a diffusion step, which disregards the laws of physics, into a physically-plausible motion $\phys{\bs{x}}^{1:H}=\mathcal{P}_\pi(\kin{\bs{x}}^{1:H})$. The projection is achieved by learning a motion imitation policy $\pi$ that controls a simulated character to mimic the denoised motion $\kin{\bs{x}}^{1:H}$ in a physics simulator. The resulting motion $\phys{\bs{x}}^{1:H}$ from the simulator is considered physically-plausible since it obeys the laws of physics.

\vspace{2mm}
\noindent\textbf{Motion Imitation Formulation.} 
The task of human motion imitation~\cite{peng2018deepmimic,yuan2020residual} can be formulated as a Markov decision process (MDP). The MDP is defined by a tuple $\mathcal{M} = (\mathcal{S}, \mathcal{A}, \mathcal{T}, R, \gamma)$ of states, actions, transition dynamics, a reward function, and a discount factor. A character agent acts in a physics simulator according to a motion imitation policy $\pi(\bs{a}^h|\bs{s}^h)$, which models the distribution of choosing an action $\bs{a}^h \in \mathcal{A}$ given the current state $\bs{s}^h \in \mathcal{S}$. The state $\bs{s}^h$ consists of the character's physical state (\eg, joint angles, velocities, positions) as well as the next pose $\kin{\bs{x}}^{h+1}$ from the input motion. Including $\kin{\bs{x}}^{h+1}$ in the state informs the policy $\pi$ to choose an action $\bs{a}^h$ that can mimic $\kin{\bs{x}}^{h+1}$ in the simulator. Starting from an initial state $\bs{s}^1$, the agent iteratively samples an action $\bs{a}^h$ from the policy $\pi$ and the simulator with transition dynamics $\mathcal{T}(\bs{s}^{h+1}|\bs{s}^h, \bs{a}^h)$ generates the next state $\bs{s}^{h+1}$, from which we can extract the simulated pose $\phys{\bs{x}}^{h+1}$. By running the policy for $H$ steps, we can obtain the physically-simulated motion $\phys{\bs{x}}^{1:H}$.

\vspace{2mm}
\noindent\textbf{Training.} During training, a reward $r^h$ is also assigned to the character based on how well the simulated motion $\phys{\bs{x}}^{1:H}$ aligns with the ground-truth motion $\gt{\bs{x}}^{1:H}$. Note that the motion imitation policy $\pi$ is trained on large motion capture datasets where high-quality ground-truth motion is available. We use reinforcement learning (RL) to learn the policy $\pi$, where the objective is to maximize the expected discounted return $J(\pi) = \mathbb{E}_{\pi}\left[\sum_{h}\gamma^h r^h\right]$ which translates to mimicking the ground-truth motion as closely as possible. We adopt a standard RL algorithm (PPO~\cite{schulman2017proximal}) to solve for the optimal policy. In the following, we will elaborate on the design of rewards, states, actions, and the policy.

\vspace{2mm}
\noindent\textbf{Rewards.} The reward function is designed to encourage the simulated motion $\phys{\bs{x}}^{1:H}$ to match the ground truth $\gt{\bs{x}}^{1:H}$. Here, we use $\;\gt{\cdot}\;$ to denote ground-truth quantities. The reward $r^h$ at each timestep consists of four sub-rewards:
\allowdisplaybreaks
\begin{align}
    r^h &= w_\texttt{p} r^h_\texttt{p} + w_\texttt{v} r^h_\texttt{v} + w_\texttt{j} r^h_\texttt{j} + w_\texttt{q} r^h_\texttt{q}\,, \\
    r^h_\texttt{p} &= \textstyle\exp\left[-\alpha_\texttt{p}\left(\sum_{j=1}^J\|\bs{o}^h_j\ominus \gt{\bs{o}}^h_j\|^2\right)\right], \\
    r^h_\texttt{v} &= \textstyle\exp\left[-\alpha_\texttt{v}\|\bs{v}^h - \gt{\bs{v}}^h\|^2\right], \\
    r^h_\texttt{j} &= \textstyle\exp\left[-\alpha_\texttt{j}\left(\sum_{j=1}^J\|\bs{p}^h_j - \gt{\bs{p}}^h_j\|^2\right)\right], \\
    r^h_\texttt{q} &= \textstyle\exp\left[-\alpha_\texttt{q}\left(\sum_{j=1}^J\|\bs{q}^h_j - \gt{\bs{q}}^h_j\|^2\right)\right].
\end{align}
where $w_\texttt{p},w_\texttt{v},w_\texttt{j},w_\texttt{q},\alpha_\texttt{p},\alpha_\texttt{v},\alpha_\texttt{j},\alpha_\texttt{q}$ are weighting factors. The pose reward $r^h_\texttt{p}$ measures the difference between the local joint rotations $\bs{o}^h_j$ and the ground truth~$\gt{\bs{o}}^h_j$, where $\ominus$ denotes the relative rotation between two rotations, and $\|\cdot\|$ computes the rotation angle.
The velocity reward $r^h_\texttt{v}$ measures the mismatch between joint velocities $\bs{v}^h$ and the ground truth $\gt{\bs{v}}^h$, which are computed via finite difference.
The joint position reward $r^h_\texttt{j}$ encourages the 3D world joint positions $\bs{p}^h_j$ to match the ground truth  $\gt{\bs{p}}^h_j$.
Finally, the joint rotation reward $r^h_\texttt{q}$ measures the difference between the global joint rotations $\bs{q}^h_j$ and the ground truth $\gt{\bs{q}}^h_j$.

\vspace{2mm}
\noindent\textbf{States.} The agent state $\bs{s}^h$ consists of the character's current physical state, the input motion's next pose $\kin{\bs{x}}^{h+1}$, and a character attribute vector $\bs{\psi}$. The character's physical state includes its joint angles, joint velocities, and rigid bodies' positions, rotations, and linear and angular velocities. For the input pose $\kin{\bs{x}}^{h+1}$, the state $\bs{s}^h$ contains the difference of $\kin{\bs{x}}^{h+1}$ \wrt the agent in joint angles as well as rigid body positions and rotations. Using the difference informs the policy about the pose residual it needs to compensate for. All the features are computed in the character's heading coordinate to ensure rotation and translation invariance. Since our character is based on the SMPL body model~\cite{loper2015smpl}, the attribute $\bs{\psi}$ includes the gender and SMPL shape parameters to allow the policy to control different characters.

\vspace{2mm}
\noindent\textbf{Actions.} We use the target joint angles of proportional derivative (PD) controllers as the action representation, which enables robust motion imitation as observed in prior work~\cite{peng2017learning,yuan2020residual}. We also add residual forces~\cite{yuan2020residual} in the action space to stabilize the character and compensate for missing contact forces required to imitate motions such as sitting.

\vspace{2mm}
\noindent\textbf{Policy.}
We use a parametrized Gaussian policy $\pi(\bs{a}^h|\bs{s}^h) = \mathcal{N}(\bs{\mu}_\theta(\bs{s}^h), \bs{\Sigma})$ where the mean action $\bs{\mu}_\theta$ is output by a simple multi-layer perceptron (MLP) network with parameters $\theta$, and $\bs{\Sigma}$ is a fixed diagonal covariance matrix.

\section{Experiments}
\label{sec:exp}
We perform experiments on two standard human motion generation tasks: text-to-motion and action-to-motion generation. In particular, our experiments are designed to answer the following questions: (1) Can PhysDiff achieve SOTA motion quality and physical plausibility? (2) Can PhysDiff be applied to different kinematic motion diffusion models to improve their motion quality and physical plausibility?  (3) How do different schedules of the physics-based projection impact motion generation performance? (4) Can PhysDiff outperform physics-based post-processing?

\vspace{1mm}
\noindent\textbf{Evaluation Metrics.}
For text-to-motion generation, we first use two standard metrics suggested by Guo \etal~\cite{guo2022generating}: \emph{FID} measures the distance between the generated and ground-truth motion distributions; \emph{R-Precision} assesses the relevancy of the generated motions to the input text. For action-to-motion generation, we replace R-Precision with an \emph{Accuracy} metric, which measures the accuracy of a trained action classifier over the generated motion. Additionally, we also use four physics-based metrics to evaluate the physical plausibility of generated motions: \emph{Penetrate} measures ground penetration; \emph{Float} measures floating; \emph{Skate} measures foot sliding; \emph{Phys-Err} is an overall physical error metric that sums the three metrics (all in \emph{mm}) together. Please refer to Appendix~\ref{sec:supp:metrics} for details.

\vspace{1mm}
\noindent\textbf{Implementation Details.}
Our model uses 50 diffusion steps with classifier-free guidance~\cite{ho2022classifier}. We test PhysDiff with two SOTA motion diffusion models, MDM~\cite{tevet2022human} and MotionDiffuse~\cite{zhang2022motiondiffuse}, as the denoiser $D$. By default, MDM is the denoiser of PhysDiff for qualitative results. We adopt IsaacGym~\cite{makoviychuk2021isaac} as the physics simulator for motion imitation. More details are provided in Appendices~\ref{sec:supp:diffusion} and \ref{sec:supp:imitation}.

\subsection{Text-to-Motion Generation}
\input{tables/humanml3d}
\input{tables/humanact12}
\input{tables/uestc}

\label{sec:exp:t2m}
\noindent\textbf{Data.}
We use the HumanML3D~\cite{guo2022generating} dataset, which is a textually annotated subset of two large-scale motion capture datasets, AMASS~\cite{mahmood2019amass} and HumanAct12~\cite{guo2020action2motion}. It contains 14,616 motions annotated with 44,970 textual descriptions.

\begin{figure*}
    \centering
    \includegraphics[width=\linewidth]{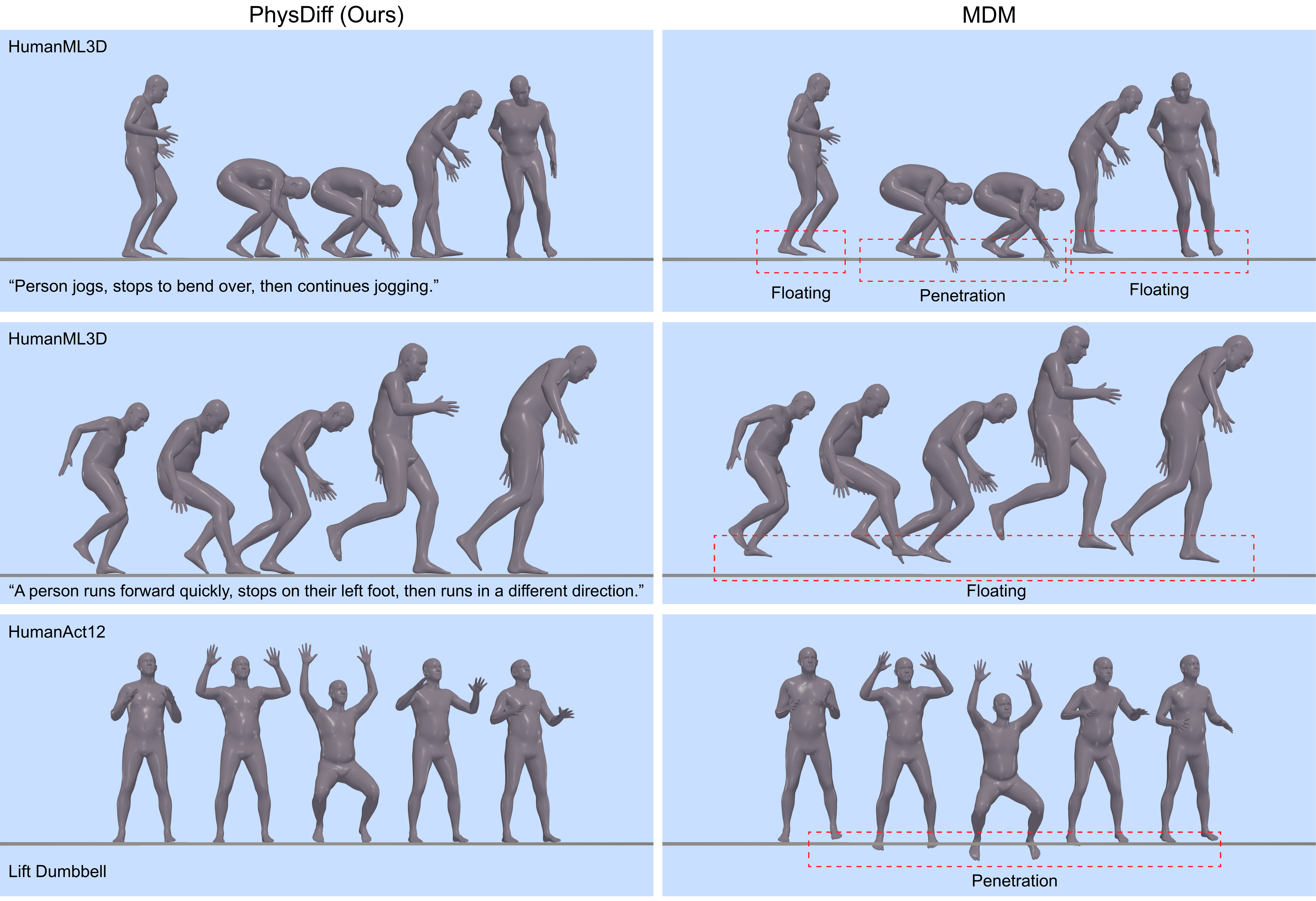}
    \vspace{-6mm}
    \caption{Visual comparison of PhysDiff against the SOTA, MDM~\cite{tevet2022human}, on HumanML3D, HumanAct12, and UESTC. PhysDiff reduces physical artifacts such as floating and penetration significantly. Please refer to the \href{https://nvlabs.github.io/PhysDiff}{project page} for more qualitative comparison.}
    \label{fig:vis-res}
    \vspace{-2mm}
\end{figure*}

\vspace{2mm}
\noindent\textbf{Results.} In Table~\ref{table:humanml3d}, we compare our method to the SOTA methods: JL2P~\cite{ahuja2019language2pose}, Text2Gesture~\cite{bhattacharya2021text2gestures},  T2M~\cite{guo2022generating}, MotionDiffuse~\cite{zhang2022motiondiffuse}, and MDM~\cite{tevet2022human}. Due to the plug-and-play nature of our method, we design two variants of PhysDiff using MotionDiffuse (MD) and MDM. PhysDiff with MDM achieves SOTA FID and also reduces Phys-Err by more than 86\% compared to MDM. Similarly, PhysDiff with MD achieves SOTA in physics-based metrics while maintaining high R-Precision and improving FID significantly. We also provide qualitative comparison in Fig.~\ref{fig:vis-res}, where we can clearly see that PhysDiff substantially reduces physical artifacts such as penetration and floating. Please also refer to the \href{https://nvlabs.github.io/PhysDiff}{project page} for more qualitative results.

\subsection{Action-to-Motion Generation}
\label{sec:exp:a2m}

\noindent\textbf{Data.}
We evaluate on two datasets: HumanAct12~\cite{guo2020action2motion}, which contains around 1200 motion clips for 12 action categories; UESTC~\cite{ji2018large}, which consists of 40 action classes, 40 subjects, and 25k samples. For both datasets, we use the sequences provided by Petrovich \etal~\cite{petrovich2021action}.

\vspace{2mm}
\noindent\textbf{Results.}
Tables~\ref{table:humanact12} and \ref{table:uestc} summarize the results on HumanAct12 and UESTC, respectively, where we compare \mbox{PhysDiff} against the SOTA methods: MDM~\cite{tevet2022human}, INR~\cite{cervantes2022implicit}, Action2Motion~\cite{guo2020action2motion}, and ACTOR~\cite{petrovich2021action}. The results show that our method achieves competitive FID on both datasets while drastically improving Phys-Err (by 78\% on HumanAct12 and 94\% on UESTC). Please refer to Fig.~\ref{fig:vis-res} and the \href{https://nvlabs.github.io/PhysDiff}{project page} for qualitative comparison, where we show that PhysDiff improves the physical plausibility of generated motions significantly.

\subsection{Schedule of Physics-Based Projection}
\label{sec:schedule}
We perform extensive experiments to analyze the schedule of the physics-based projection, \ie, at which timesteps we perform the projection in the diffusion process.

\begin{figure}[t]
    \centering
    \includegraphics[width=\linewidth]{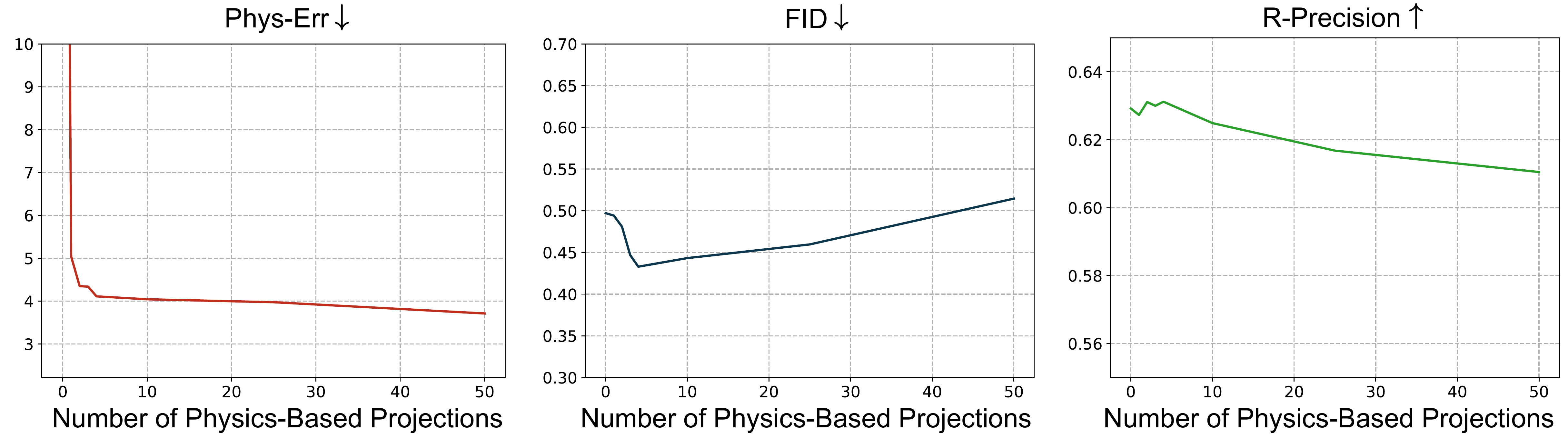}
    \vspace{-6mm}
    \caption{Effect of varying the number of physics-based projection steps for text-to-motion generation on HumanML3D~\cite{guo2022generating}.}
    \label{fig:schedule}
    \vspace{-3mm}
\end{figure}

\vspace{2mm}
\noindent\textbf{Number of Projection Steps.}
Since the physics-based projection is relatively expensive to compute, we first investigate whether we can reduce the number of projection steps without sacrificing performance. To this end, we vary the number of projection steps performed during diffusion from 50 to 0, where the projection steps are gradually removed from earlier timesteps and applied consecutively. We plot the curves of FID, R-Precision, and Phys-Err in Fig.~\ref{fig:schedule}. As can be seen, Phys-Err keeps decreasing with more physics-based projection steps, which indicates more projection steps always help improve the physical plausibility of PhysDiff. Interestingly, both the FID and R-Precision first improve (FID decreases and R-Precision increases) and then deteriorate when increasing the number of projection steps. This suggests that there is a trade-off between physical plausibility and motion quality when more projection steps are performed at the early diffusion steps. We hypothesize that motions generated at the early diffusion steps are denoised to the mean motion of the dataset (with little body movement) and are often not physically-plausible. As a result, performing the physics-based projection at these early steps can push the generated motion away from the data distribution, thus hindering the diffusion process.

\input{tables/abl_schedule}

\vspace{2mm}
\noindent\textbf{Placement of Projections Steps.}
Fig.~\ref{fig:schedule} indicates that four physics-based projection steps yield a good trade-off between physical plausibility and motion quality. Next, we investigate the best placement of these projection steps in the diffusion process. We compare three groups of schedules: (1) \emph{Uniform $N$}, which spreads the $N$ projection steps evenly across the diffusion timesteps \ie, for 50 diffusion steps and $N=4$, the projection steps are performed at $t \in \{0, 15, 30, 45\}$; (2) \emph{Start $M$, End $N$}, which places $M$ consecutive projection steps at the beginning of the diffusion process and $N$ projection steps at the end; (3) \emph{End $N$, Space $S$}, which places $N$ projections steps with time spacing $S$ at the end of the diffusion process (\eg, for $N=4,S=3$, the projections steps are performed at $t \in \{0, 3, 6, 9\}$). We summarize the results in Table~\ref{table:abl_schedule}. We can see that the schedule \emph{Start $M$, End $N$} has inferior FID and R-Precision since more physics-based projection steps are performed at early diffusion steps, which is consistent with our findings in Fig.~\ref{fig:schedule}. The schedule \emph{Uniform $N$} works better in terms of FID and R-Precision but has worse Phys-Err. This is likely because too many non-physics-based diffusion steps between the physics-based projections undo the effect of the projection and reintroduce physical errors. This is also consistent with \emph{End $4$, Space $3$} being worse than \emph{End $4$, Space $1$} since the former has more diffusion steps between the physics-based projections. Hence, the results suggest that it is better to schedule the physics-based projection steps consecutively toward the end. This guides us to use \emph{End $4$, Space $1$} for baseline comparison.

\vspace{2mm}
\noindent\textbf{Inference Time.}
Due to the use of physics simulation, PhysDiff is 2.5x slower than MDM (51.6s vs.\,19.6s) to generate a single motion, where both methods use 1000 diffusion steps. The gap closes when increasing the batch size to 256 where PhysDiff is only 1.7x slower (280.3s vs.\,471.3s), as the physics simulator benefits more from parallelization.

\begin{figure}[t]
    \centering
    \includegraphics[width=\linewidth]{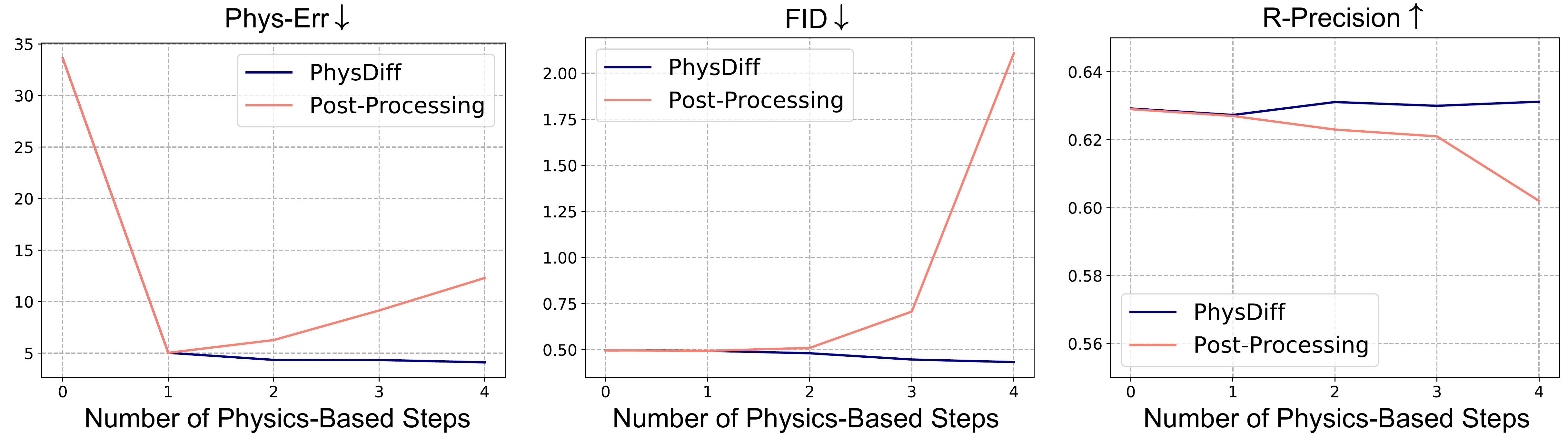}
    \vspace{-6mm}
    \caption{Comparison with post-processing, \ie, applying physics-based projections after the diffusion process.}
    \label{fig:vs_postproc}
    \vspace{-2mm}
\end{figure}

\subsection{Comparing against Post-Processing}
\label{sec:vs_postproc}
To demonstrate the synergy between physics and diffusion, we compare PhysDiff against a post-processing baseline that applies one or more physics-based projection steps to the final kinematic motion from diffusion. As shown in Fig.~\ref{fig:vs_postproc}, multiple post-processing steps cannot enhance motion quality or physical plausibility; instead, they deteriorate them. This is because the final kinematic motion may be too physically implausible for the physics to imitate, \eg, the human may lose balance due to wrong gaits. Repeatedly imitating these implausible motions could amplify the problem and lead to unstable simulation. PhysDiff overcomes this issue by iteratively applying diffusion and physics to recover from bad simulation states and move closer to the data distribution.

\section{Conclusion and Future Work}
In this paper, we proposed a novel physics-guided motion diffusion model (PhysDiff) which instills the laws of physics into the diffusion process to generate physically-plausible human motions. To achieve this, we proposed a physics-based motion projection module that uses motion imitation in physics simulation to enforce physical constraints.
Our approach is agnostic to the denoising network and can be used to improve SOTA motion diffusion models without retraining. Experiments on large-scale motion data demonstrate that PhysDiff achieves SOTA motion quality and substantially improves physical plausibility.

Due to physics simulation, the inference speed of PhysDiff can be two-to-three times slower than SOTA models. Future work could speed up the model with a faster physics simulator or improve the physics-based projection to reduce the number of required projection steps.

{\small
\bibliographystyle{ieee_fullname}
\bibliography{references}
}

\clearpage
\appendix
\section{Details of Evaluation Metrics}
\label{sec:supp:metrics}
We use the open source \href{https://github.com/GuyTevet/motion-diffusion-model}{code}\footnote{\url{https://github.com/GuyTevet/motion-diffusion-model}} of MDM~\cite{tevet2022human} to compute the motion-based metrics: \emph{FID}, \emph{R-Precision}, and \emph{Accuracy}. The physics-based metrics are implemented as follows. For ground penetration (\emph{Penetrate}), we compute the distance between the ground and the lowest body mesh vertex below the ground. For floating (\emph{Float}), we compute the distance between the ground and the lowest body mesh vertex above the ground. For both \emph{Penetrate} and \emph{Float}, we have a tolerance of 5 mm to account for geometry approximation. For foot sliding (\emph{Skate}), we find foot joints that contact the ground in two adjacent frames and compute their average horizontal displacement within the frames. The overall physics error metric \emph{Phys-Err} is the sum of \emph{Penetrate}, \emph{Float}, and \emph{Skate}.

\section{Details of Motion Diffusion}
\label{sec:supp:diffusion}

As mentioned in the main paper, we tested PhysDiff with two state-of-the-art denoiser networks, MDM~\cite{tevet2022human} and MotionDiffuse~\cite{zhang2022motiondiffuse} and showed that PhysDiff can improve both of them. We directly use the pretrained models in their codebase. Please refer to their paper and code for additional details.

For diffusion sampling, we use 50 timesteps with $\eta = 0$. We also use classifier-free guidance with the guidance coefficient set to 2.5. For text-to-motion generation on HumanML3D~\cite{guo2022generating}, the data is represented by a 263-dim vector that consists of 3D joint positions, rotations, and velocities, following Guo \etal~\cite{guo2022generating}. To perform the physics-based motion projection, we first convert the 3D joint positions into joint angles of the SMPL model~\cite{loper2015smpl} using inverse kinematics and then apply physics-based motion imitation. For action-to-motion generation, the data is represented by joint rotations, so no inverse kinematics is required.

\vspace{2mm}
\noindent\textbf{Policy Training.}
The motion imitation policy uses a three-layer MLP with hidden dimensions (1024, 1024, 512) and ReLU activations. The elements of the policy's diagonal covariance matrix $\bs{\Sigma}$ are set to 0.173. We also normalize the policy's input state using a running estimate of the mean and variance of the state. We train the policy using the AMASS~\cite{mahmood2019amass} human motion database. Since HumanML3D is a text-annotated version of AMASS, we use the same training split as HumanML3D and do not use additional data for fair comparison. We created 8192 parallel simulation environments in IsaacGym to collect training samples. Each RL episode has a horizon of 32 frames. We train the policy for 4000 epochs where each epoch collects 262,144 samples from running all environments for an episode. The reward weights ($w_\texttt{p},w_\texttt{v},w_\texttt{j},w_\texttt{q}$) are set to (0.6, 0.1, 0.2, 0.1), and the reward parameters ($\alpha_\texttt{p},\alpha_\texttt{v},\alpha_\texttt{j},\alpha_\texttt{q}$) are set to (60, 0.2, 100, 40). Proximal policy optimization (PPO~\cite{schulman2017proximal}) is used to train the policy. The clipping coefficient $\epsilon$ in PPO is set to 0.2. The discount factor $\gamma$ for the Markov decision process (MDP) is set to 0.99. We also use the generalized advantage estimator GAE($\lambda$)~\cite{schulman2015high} to estimate the advantage for policy gradient, and the GAE coefficient $\lambda$ is 0.95. At the end of each epoch, we update the policy by iterating over the samples for 6 mini-epochs with a mini-batch size of 512. The update is performed via Adam~\cite{kingma2014adam} with a base learning rate of $2\times 10^{-5}$. We clip the gradient if its norm is larger than 50.

\begin{table}[h]
\footnotesize
\centering
\begin{tabular}{lc}
\toprule
Parameter & Value\\ \midrule
Num. of simulation environments & 8192 \\
Episode horizon & 32 \\
Num. of epochs & 4000 \\
Num. of mini-epochs & 6 \\
Learning rate & $2\times 10^{-5}$\\
PPO clip $\epsilon$ & 0.2 \\
Discount factor $\gamma$ & 0.99 \\
GAE coefficient $\lambda$ & 0.95 \\
Reward weights ($w_\texttt{p},w_\texttt{v},w_\texttt{j},w_\texttt{q}$) & (0.6, 0.1, 0.2, 0.1) \\
Reward parameters ($\alpha_\texttt{p},\alpha_\texttt{v},\alpha_\texttt{j},\alpha_\texttt{q}$) & (60, 0.2, 100, 40) \\
Elements of diagonal covariance $\bs{\Sigma}$ & 0.173 \\
\bottomrule
\end{tabular}
\vspace{1mm}
\caption{Hyperparameters for physics-based motion imitation.}
\label{table:hyper-im}
\end{table}

\section{Details of Physics-Based Motion Imitation}
\label{sec:supp:imitation}
\noindent\textbf{Physics Simulation and Character.}
We use IsaacGym~\cite{makoviychuk2021isaac} as our physics simulator for its ability to perform massively parallel simulation on GPUs. The simulation runs at 60Hz while the policy controls the character at 30Hz. The character is automatically created from SMPL parameters following the approach in \mbox{SimPoE}~\cite{yuan2021simpoe}.

\end{document}

%% file: def.tex
\newcommand{\bs}[1]{\boldsymbol{#1}}
\newcommand{\phys}[1]{\hat{#1}}
\newcommand{\kin}[1]{\tilde{#1}}
\newcommand{\gt}[1]{\bar{#1}}

%% file: tables/humanml3d.tex
\setlength{\tabcolsep}{3pt}
\begin{table}[t]
\footnotesize
\centering
\resizebox{\linewidth}{!}{ 
\begin{tabular}{lcccccc}
\toprule
Method       & FID\,$\downarrow$ & R-Precision\,$\uparrow$ & Penetrate\,$\downarrow$ & Float\,$\downarrow$ & Skate\,$\downarrow$ & Phys-Err\,$\downarrow$\\ \midrule
J2LP~\cite{ahuja2019language2pose}                & 11.020         & 0.486          & -              & -              & -              & -              \\
Text2Gesture~\cite{bhattacharya2021text2gestures} & 7.664          & 0.345          & -              & -              & -              & -              \\
T2M~\cite{guo2022generating}                      & 1.067          & 0.740 & 11.897         & 7.779          & 2.908          & 22.584         \\
MDM~\cite{tevet2022human}                         & 0.544          & 0.611          & 11.291         & 18.876         & 1.406          & 31.572         \\
MotionDiffuse (MD)~\cite{zhang2022motiondiffuse} & 0.630  & \textbf{0.782} &   20.278       &  6.450         & 3.925          & 30.652              \\
\midrule
PhysDiff w/ MD (Ours)    & 0.551 & 0.780     & \textbf{0.898}        & \textbf{1.368} & \textbf{0.423} & \textbf{2.690} \\
PhysDiff w/ MDM (Ours)                                      & \textbf{0.433} & 0.631          & 0.998 & 2.601 & 0.512 & 4.111 \\
\bottomrule
\end{tabular}
}
\vspace{-2.5mm}
\caption{Text-to-motion results on HumanML3D~\cite{guo2022generating}.}
\label{table:humanml3d}
\vspace{-3mm}
\end{table}

%% file: tables/humanact12.tex
\setlength{\tabcolsep}{3pt}
\begin{table}[t]
\footnotesize
\centering
\resizebox{\linewidth}{!}{ 
\begin{tabular}{lcccccc}
\toprule
Method       & FID\,$\downarrow$ & Accuracy\,$\uparrow$ & Penetrate\,$\downarrow$ & Float\,$\downarrow$ & Skate\,$\downarrow$ & Phys-Err\,$\downarrow$\\ \midrule
Action2Motion~\cite{guo2020action2motion} & 0.338          & 0.917          & -              & -              & -              & -              \\
ACTOR~\cite{petrovich2021action}          & 0.120          & 0.955          & 8.939          & 14.479         & 2.277          & 25.695          \\
INR~\cite{cervantes2022implicit}          & \textbf{0.088} & 0.973          & 7.055          & 13.212         & 0.928          & 22.096         \\
MDM~\cite{tevet2022human}                 & 0.100          & \textbf{0.990} & 5.600          & 6.703          & 1.075          & 13.377         \\ \midrule
PhysDiff w/ MDM (Ours)                               & 0.096          & 0.983          & \textbf{0.689} & \textbf{2.002} & \textbf{0.159} & \textbf{2.850} \\
\bottomrule
\end{tabular}
}
\vspace{-2.5mm}
\caption{Action-to-motion results on HumanAct12~\cite{guo2020action2motion}.}
\label{table:humanact12}
\vspace{-3mm}
\end{table}

%% file: tables/uestc.tex
\setlength{\tabcolsep}{3pt}
\begin{table}[t]
\footnotesize
\centering
\resizebox{\linewidth}{!}{ 
\begin{tabular}{lcccccc}
\toprule
Method       & FID\,$\downarrow$ & Accuracy\,$\uparrow$ & Penetrate\,$\downarrow$ & Float\,$\downarrow$ & Skate\,$\downarrow$ & Phys-Err\,$\downarrow$\\ \midrule
ACTOR~\cite{petrovich2021action}  & 23.43          & 0.911          & 8.441          & 9.737          & 1.073          & 19.251         \\
INR~\cite{cervantes2022implicit}  & 15.00          & 0.941          & 5.999          & 4.633          & 0.741          & 11.373         \\
MDM~\cite{tevet2022human}         & \textbf{12.81} & 0.950          & 13.077         & 13.912         & 1.383          & 28.371         \\ \midrule
PhysDiff w/ MDM (Ours) & 13.27               & \textbf{0.956} & \textbf{0.874} & \textbf{0.201} & \textbf{0.389} & \textbf{1.463} \\
\bottomrule
\end{tabular}
}
\vspace{-2.5mm}
\caption{Action-to-motion results on UESTC~\cite{ji2018large}.}
\label{table:uestc}
\vspace{-4mm}
\end{table}

%% file: tables/abl_schedule.tex
\setlength{\tabcolsep}{3pt}
\begin{table}[t]
\footnotesize
\centering
\resizebox{\linewidth}{!}{ 
\begin{tabular}{l@{\hskip 5mm}cccccc}
\toprule
Schedule       & FID\,$\downarrow$ & R-Precision\,$\uparrow$ & Penetrate\,$\downarrow$ & Float\,$\downarrow$ & Skate\,$\downarrow$ & Phys-Err\,$\downarrow$\\ \midrule
Uniform 4      & 0.473          & 0.630          & 0.979          & 3.479          & 0.463          & 4.921          \\\midrule
Start 3, End 1 & 0.510          & 0.623          & \textbf{0.918}          & 3.173          & \textbf{0.459}          & 4.550          \\
Start 2, End 2 & 0.503          & 0.623          & \textbf{0.918}          & 2.723          & 0.492          & 4.133          \\\midrule
End 4, Space 3  & 0.469          & 0.630          & 0.990          & 3.226          & 0.473          & 4.689          \\
End 4, Space 2  & 0.469          & 0.630          & 0.990          & 3.004          & 0.476          & 4.470          \\
End 4, Space 1  & \textbf{0.433} & \textbf{0.631} & 0.998 & \textbf{2.601} & 0.512 & \textbf{4.111}\\
\bottomrule
\end{tabular}
}
\vspace{-2.5mm}
\caption{Projection schedule comparison on HumanML3D~\cite{guo2022generating}.}
\label{table:abl_schedule}
\vspace{-2mm}
\end{table}

%% file: arxiv.bbl
\begin{thebibliography}{10}\itemsep=-1pt

\bibitem{ahuja2019language2pose}
Chaitanya Ahuja and Louis-Philippe Morency.
\newblock Language2pose: Natural language grounded pose forecasting.
\newblock In {\em 2019 International Conference on 3D Vision (3DV)}, pages
  719--728. IEEE, 2019.

\bibitem{aksan2019structured}
Emre Aksan, Manuel Kaufmann, and Otmar Hilliges.
\newblock Structured prediction helps 3d human motion modelling.
\newblock In {\em Proceedings of the IEEE International Conference on Computer
  Vision}, pages 7144--7153, 2019.

\bibitem{aliakbarian2020stochastic}
Sadegh Aliakbarian, Fatemeh~Sadat Saleh, Mathieu Salzmann, Lars Petersson, and
  Stephen Gould.
\newblock A stochastic conditioning scheme for diverse human motion prediction.
\newblock In {\em Proceedings of the IEEE/CVF Conference on Computer Vision and
  Pattern Recognition}, pages 5223--5232, 2020.

\bibitem{aneja2021contrastive}
Jyoti Aneja, Alex Schwing, Jan Kautz, and Arash Vahdat.
\newblock A contrastive learning approach for training variational autoencoder
  priors.
\newblock {\em Advances in Neural Information Processing Systems}, 34:480--493,
  2021.

\bibitem{bao2022estimating}
Fan Bao, Chongxuan Li, Jiacheng Sun, Jun Zhu, and Bo Zhang.
\newblock Estimating the optimal covariance with imperfect mean in diffusion
  probabilistic models.
\newblock In {\em International Conference on Machine Learning}, 2022.

\bibitem{bao2022analytic}
Fan Bao, Chongxuan Li, Jun Zhu, and Bo Zhang.
\newblock {Analytic-DPM}: {A}n analytic estimate of the optimal reverse
  variance in diffusion probabilistic models.
\newblock In {\em International Conference on Learning Representations}, 2022.

\bibitem{barsoum2018hp}
Emad Barsoum, John Kender, and Zicheng Liu.
\newblock Hp-gan: Probabilistic 3d human motion prediction via gan.
\newblock In {\em Proceedings of the IEEE Conference on Computer Vision and
  Pattern Recognition Workshops}, pages 1418--1427, 2018.

\bibitem{bergamin2019drecon}
Kevin Bergamin, Simon Clavet, Daniel Holden, and James~Richard Forbes.
\newblock Drecon: data-driven responsive control of physics-based characters.
\newblock {\em ACM Transactions on Graphics (TOG)}, 38(6):1--11, 2019.

\bibitem{bhattacharya2021text2gestures}
Uttaran Bhattacharya, Nicholas Rewkowski, Abhishek Banerjee, Pooja Guhan,
  Aniket Bera, and Dinesh Manocha.
\newblock Text2gestures: A transformer-based network for generating emotive
  body gestures for virtual agents.
\newblock In {\em 2021 IEEE Virtual Reality and 3D User Interfaces (VR)}, pages
  1--10. IEEE, 2021.

\bibitem{cervantes2022implicit}
Pablo Cervantes, Yusuke Sekikawa, Ikuro Sato, and Koichi Shinoda.
\newblock Implicit neural representations for variable length human motion
  generation.
\newblock In {\em Proceedings of the European Conference on Computer Vision
  (ECCV)}, pages 356--372. Springer, 2022.

\bibitem{choi2021ilvr}
Jooyoung Choi, Sungwon Kim, Yonghyun Jeong, Youngjune Gwon, and Sungroh Yoon.
\newblock {ILVR}: Conditioning method for denoising diffusion probabilistic
  models.
\newblock {\em arXiv preprint arXiv:2108.02938}, August 2021.

\bibitem{chung2021come}
Hyungjin Chung, Byeongsu Sim, and Jong~Chul Ye.
\newblock {Come-Closer-Diffuse-Faster}: Accelerating conditional diffusion
  models for inverse problems through stochastic contraction.
\newblock {\em arXiv preprint arXiv:2112.05146}, December 2021.

\bibitem{dhariwal2021diffusion}
Prafulla Dhariwal and Alexander~Quinn Nichol.
\newblock Diffusion models beat {GAN}s on image synthesis.
\newblock In {\em Advances in Neural Information Processing Systems}, 2021.

\bibitem{dockhorn2022genie}
Tim Dockhorn, Arash Vahdat, and Karsten Kreis.
\newblock {{GENIE}}: {H}igher-order denoising diffusion solvers.
\newblock In {\em Advances in Neural Information Processing Systems}, 2022.

\bibitem{dockhorn2021score}
Tim Dockhorn, Arash Vahdat, and Karsten Kreis.
\newblock Score-based generative modeling with critically-damped {L}angevin
  diffusion.
\newblock In {\em International Conference on Learning Representations}, 2022.

\bibitem{efron2011tweedie}
Bradley Efron.
\newblock Tweedie’s formula and selection bias.
\newblock {\em Journal of the American Statistical Association},
  106(496):1602--1614, 2011.

\bibitem{fragkiadaki2015recurrent}
Katerina Fragkiadaki, Sergey Levine, Panna Felsen, and Jitendra Malik.
\newblock Recurrent network models for human dynamics.
\newblock In {\em Proceedings of the IEEE International Conference on Computer
  Vision}, pages 4346--4354, 2015.

\bibitem{ghosh2017learning}
Partha Ghosh, Jie Song, Emre Aksan, and Otmar Hilliges.
\newblock Learning human motion models for long-term predictions.
\newblock In {\em 2017 International Conference on 3D Vision (3DV)}, pages
  458--466. IEEE, 2017.

\bibitem{goodfellow2014generative}
Ian Goodfellow, Jean Pouget-Abadie, Mehdi Mirza, Bing Xu, David Warde-Farley,
  Sherjil Ozair, Aaron Courville, and Yoshua Bengio.
\newblock Generative adversarial nets.
\newblock In {\em Advances in Neural Information Processing Systems}, pages
  2672--2680, 2014.

\bibitem{gopalakrishnan2019neural}
Anand Gopalakrishnan, Ankur Mali, Dan Kifer, Lee Giles, and Alexander~G
  Ororbia.
\newblock A neural temporal model for human motion prediction.
\newblock In {\em Proceedings of the IEEE Conference on Computer Vision and
  Pattern Recognition}, pages 12116--12125, 2019.

\bibitem{grenander1994representations}
Ulf Grenander and Michael~I Miller.
\newblock Representations of knowledge in complex systems.
\newblock {\em Journal of the Royal Statistical Society: Series B
  (Methodological)}, 56(4):549--581, 1994.

\bibitem{guo2022generating}
Chuan Guo, Shihao Zou, Xinxin Zuo, Sen Wang, Wei Ji, Xingyu Li, and Li Cheng.
\newblock Generating diverse and natural 3d human motions from text.
\newblock In {\em Proceedings of the IEEE/CVF Conference on Computer Vision and
  Pattern Recognition}, pages 5152--5161, 2022.

\bibitem{guo2020action2motion}
Chuan Guo, Xinxin Zuo, Sen Wang, Shihao Zou, Qingyao Sun, Annan Deng, Minglun
  Gong, and Li Cheng.
\newblock Action2motion: Conditioned generation of 3d human motions.
\newblock In {\em Proceedings of the 28th ACM International Conference on
  Multimedia}, pages 2021--2029, 2020.

\bibitem{he2022nemf}
Chengan He, Jun Saito, James Zachary, Holly Rushmeier, and Yi Zhou.
\newblock Nemf: Neural motion fields for kinematic animation.
\newblock {\em arXiv preprint arXiv:2206.03287}, 2022.

\bibitem{ho2020denoising}
Jonathan Ho, Ajay Jain, and Pieter Abbeel.
\newblock Denoising diffusion probabilistic models.
\newblock In {\em Advances in Neural Information Processing Systems}, 2020.

\bibitem{ho2022classifier}
Jonathan Ho and Tim Salimans.
\newblock Classifier-free diffusion guidance.
\newblock In {\em NeurIPS 2021 Workshop on Deep Generative Models and
  Downstream Applications}, 2021.

\bibitem{hoyet2012push}
Ludovic Hoyet, Rachel McDonnell, and Carol O'Sullivan.
\newblock Push it real: Perceiving causality in virtual interactions.
\newblock {\em ACM Transactions on Graphics (TOG)}, 31(4):1--9, 2012.

\bibitem{isogawa2020optical}
Mariko Isogawa, Ye Yuan, Matthew O'Toole, and Kris~M Kitani.
\newblock Optical non-line-of-sight physics-based 3d human pose estimation.
\newblock In {\em Proceedings of the IEEE/CVF Conference on Computer Vision and
  Pattern Recognition}, pages 7013--7022, 2020.

\bibitem{ji2018large}
Yanli Ji, Feixiang Xu, Yang Yang, Fumin Shen, Heng~Tao Shen, and Wei-Shi Zheng.
\newblock A large-scale rgb-d database for arbitrary-view human action
  recognition.
\newblock In {\em Proceedings of the 26th ACM international Conference on
  Multimedia}, pages 1510--1518, 2018.

\bibitem{karras2022elucidating}
Tero Karras, Miika Aittala, Timo Aila, and Samuli Laine.
\newblock Elucidating the design space of diffusion-based generative models.
\newblock In {\em Advances in Neural Information Processing Systems}, 2022.

\bibitem{kawar2022denoising}
Bahjat Kawar, Michael Elad, Stefano Ermon, and Jiaming Song.
\newblock Denoising diffusion restoration models.
\newblock In {\em Advances in Neural Information Processing Systems}, 2022.

\bibitem{kawar2021snips}
Bahjat Kawar, Gregory Vaksman, and Michael Elad.
\newblock {SNIPS}: Solving noisy inverse problems stochastically.
\newblock {\em arXiv preprint arXiv:2105.14951}, May 2021.

\bibitem{kingma2021variational}
Diederik Kingma, Tim Salimans, Ben Poole, and Jonathan Ho.
\newblock Variational diffusion models.
\newblock {\em Advances in Neural Information Processing Systems},
  34:21696--21707, 2021.

\bibitem{kingma2014adam}
Diederik~P Kingma and Jimmy Ba.
\newblock Adam: A method for stochastic optimization.
\newblock {\em arXiv preprint arXiv:1412.6980}, 2014.

\bibitem{kingma2013auto}
Diederik~P Kingma and Max Welling.
\newblock Auto-encoding variational bayes.
\newblock {\em arXiv preprint arXiv:1312.6114}, 2013.

\bibitem{kong2020diffwave}
Zhifeng Kong, Wei Ping, Jiaji Huang, Kexin Zhao, and Bryan Catanzaro.
\newblock {DiffWave}: {A} versatile diffusion model for audio synthesis.
\newblock In {\em International Conference on Learning Representations}, 2021.

\bibitem{li2022danceformer}
Buyu Li, Yongchi Zhao, Shi Zhelun, and Lu Sheng.
\newblock Danceformer: Music conditioned 3d dance generation with parametric
  motion transformer.
\newblock In {\em Proceedings of the AAAI Conference on Artificial
  Intelligence}, volume~36, pages 1272--1279, 2022.

\bibitem{li2021ai}
Ruilong Li, Shan Yang, David~A Ross, and Angjoo Kanazawa.
\newblock Ai choreographer: Music conditioned 3d dance generation with aist++.
\newblock In {\em Proceedings of the IEEE/CVF International Conference on
  Computer Vision}, pages 13401--13412, 2021.

\bibitem{li2017auto}
Zimo Li, Yi Zhou, Shuangjiu Xiao, Chong He, Zeng Huang, and Hao Li.
\newblock Auto-conditioned recurrent networks for extended complex human motion
  synthesis.
\newblock {\em arXiv preprint arXiv:1707.05363}, 2017.

\bibitem{liu2017learning}
Libin Liu and Jessica Hodgins.
\newblock Learning to schedule control fragments for physics-based characters
  using deep q-learning.
\newblock {\em ACM Transactions on Graphics (TOG)}, 36(3):29, 2017.

\bibitem{liu2018learning}
Libin Liu and Jessica Hodgins.
\newblock Learning basketball dribbling skills using trajectory optimization
  and deep reinforcement learning.
\newblock {\em ACM Transactions on Graphics (TOG)}, 37(4):1--14, 2018.

\bibitem{liu2022pseudo}
Luping Liu, Yi Ren, Zhijie Lin, and Zhou Zhao.
\newblock Pseudo numerical methods for diffusion models on manifolds.
\newblock In {\em International Conference on Learning Representations}, 2022.

\bibitem{loper2015smpl}
Matthew Loper, Naureen Mahmood, Javier Romero, Gerard Pons-Moll, and Michael~J
  Black.
\newblock Smpl: A skinned multi-person linear model.
\newblock {\em ACM transactions on graphics (TOG)}, 34(6):1--16, 2015.

\bibitem{lu2022dpm}
Cheng Lu, Yuhao Zhou, Fan Bao, Jianfei Chen, Chongxuan Li, and Jun Zhu.
\newblock {DPM-Solver}: {A} fast {ODE} solver for diffusion probabilistic model
  sampling in around 10 steps.
\newblock In {\em Advances in Neural Information Processing Systems}, 2022.

\bibitem{lugmayr2022repaint}
Andreas Lugmayr, Martin Danelljan, Andres Romero, Fisher Yu, Radu Timofte, and
  Luc Van~Gool.
\newblock {RePaint}: {I}npainting using denoising diffusion probabilistic
  models.
\newblock In {\em Proceedings of the IEEE/CVF Conference on Computer Vision and
  Pattern Recognition}, 2022.

\bibitem{luo2021diffusion}
Shitong Luo and Wei Hu.
\newblock Diffusion probabilistic models for {3D} point cloud generation.
\newblock In {\em Proceedings of the IEEE/CVF Conference on Computer Vision and
  Pattern Recognition}, 2021.

\bibitem{luo2021dynamics}
Zhengyi Luo, Ryo Hachiuma, Ye Yuan, and Kris Kitani.
\newblock Dynamics-regulated kinematic policy for egocentric pose estimation.
\newblock In {\em Advances in Neural Information Processing Systems}, 2021.

\bibitem{luo2022embodied}
Zhengyi Luo, Shun Iwase, Ye Yuan, and Kris Kitani.
\newblock Embodied scene-aware human pose estimation.
\newblock In {\em Advances in Neural Information Processing Systems}, 2022.

\bibitem{mahmood2019amass}
Naureen Mahmood, Nima Ghorbani, Nikolaus~F Troje, Gerard Pons-Moll, and
  Michael~J Black.
\newblock Amass: Archive of motion capture as surface shapes.
\newblock In {\em Proceedings of the IEEE/CVF international conference on
  computer vision}, pages 5442--5451, 2019.

\bibitem{makoviychuk2021isaac}
Viktor Makoviychuk, Lukasz Wawrzyniak, Yunrong Guo, Michelle Lu, Kier Storey,
  Miles Macklin, David Hoeller, Nikita Rudin, Arthur Allshire, Ankur Handa,
  et~al.
\newblock Isaac gym: High performance gpu based physics simulation for robot
  learning.
\newblock In {\em Thirty-fifth Conference on Neural Information Processing
  Systems Datasets and Benchmarks Track}, 2021.

\bibitem{mao2019learning}
Wei Mao, Miaomiao Liu, Mathieu Salzmann, and Hongdong Li.
\newblock Learning trajectory dependencies for human motion prediction.
\newblock In {\em Proceedings of the IEEE International Conference on Computer
  Vision}, pages 9489--9497, 2019.

\bibitem{martinez2017human}
Julieta Martinez, Michael~J Black, and Javier Romero.
\newblock On human motion prediction using recurrent neural networks.
\newblock In {\em Proceedings of the IEEE Conference on Computer Vision and
  Pattern Recognition}, pages 2891--2900, 2017.

\bibitem{meng2021sdedit}
Chenlin Meng, Yutong He, Yang Song, Jiaming Song, Jiajun Wu, Jun-Yan Zhu, and
  Stefano Ermon.
\newblock {SDEdit}: {G}uided image synthesis and editing with stochastic
  differential equations.
\newblock In {\em International Conference on Learning Representations}, 2022.

\bibitem{merel2017learning}
Josh Merel, Yuval Tassa, Sriram Srinivasan, Jay Lemmon, Ziyu Wang, Greg Wayne,
  and Nicolas Heess.
\newblock Learning human behaviors from motion capture by adversarial
  imitation.
\newblock {\em arXiv preprint arXiv:1707.02201}, 2017.

\bibitem{nie2022DiffPure}
Weili Nie, Brandon Guo, Yujia Huang, Chaowei Xiao, Arash Vahdat, and Anima
  Anandkumar.
\newblock Diffusion models for adversarial purification.
\newblock In {\em International Conference on Machine Learning}, 2022.

\bibitem{park2019learning}
Soohwan Park, Hoseok Ryu, Seyoung Lee, Sunmin Lee, and Jehee Lee.
\newblock Learning predict-and-simulate policies from unorganized human motion
  data.
\newblock {\em ACM Transactions on Graphics (TOG)}, 38(6):1--11, 2019.

\bibitem{pavllo2018quaternet}
Dario Pavllo, David Grangier, and Michael Auli.
\newblock Quaternet: A quaternion-based recurrent model for human motion.
\newblock {\em arXiv preprint arXiv:1805.06485}, 2018.

\bibitem{peng2018deepmimic}
Xue~Bin Peng, Pieter Abbeel, Sergey Levine, and Michiel van~de Panne.
\newblock Deepmimic: Example-guided deep reinforcement learning of
  physics-based character skills.
\newblock {\em ACM Transactions on Graphics (TOG)}, 37(4):1--14, 2018.

\bibitem{peng2018sfv}
Xue~Bin Peng, Angjoo Kanazawa, Jitendra Malik, Pieter Abbeel, and Sergey
  Levine.
\newblock Sfv: Reinforcement learning of physical skills from videos.
\newblock {\em ACM Transactions on Graphics (TOG)}, 37(6):1--14, 2018.

\bibitem{peng2017learning}
Xue~Bin Peng and Michiel van~de Panne.
\newblock Learning locomotion skills using deeprl: Does the choice of action
  space matter?
\newblock In {\em Proceedings of the ACM SIGGRAPH/Eurographics Symposium on
  Computer Animation}, pages 1--13, 2017.

\bibitem{petrovich2021action}
Mathis Petrovich, Michael~J Black, and G{\"u}l Varol.
\newblock Action-conditioned 3d human motion synthesis with transformer vae.
\newblock In {\em Proceedings of the IEEE/CVF International Conference on
  Computer Vision}, pages 10985--10995, 2021.

\bibitem{petrovich22temos}
Mathis Petrovich, Michael~J. Black, and G{\"u}l Varol.
\newblock {TEMOS}: Generating diverse human motions from textual descriptions.
\newblock In {\em Proceedings of the European Conference on Computer Vision
  (ECCV)}, 2022.

\bibitem{ramesh2022hierarchical}
Aditya Ramesh, Prafulla Dhariwal, Alex Nichol, Casey Chu, and Mark Chen.
\newblock Hierarchical text-conditional image generation with {CLIP} latents.
\newblock {\em arXiv preprint arXiv:2204.06125}, 2022.

\bibitem{reitsma2003perceptual}
Paul~SA Reitsma and Nancy~S Pollard.
\newblock Perceptual metrics for character animation: sensitivity to errors in
  ballistic motion.
\newblock In {\em ACM SIGGRAPH 2003 Papers}, pages 537--542, 2003.

\bibitem{rempe2020}
Davis Rempe, Leonidas~J. Guibas, Aaron Hertzmann, Bryan Russell, Ruben
  Villegas, and Jimei Yang.
\newblock Contact and human dynamics from monocular video.
\newblock In {\em Proceedings of the European Conference on Computer Vision
  (ECCV)}, 2020.

\bibitem{ren2022diffusion}
Zhiyuan Ren, Zhihong Pan, Xin Zhou, and Le Kang.
\newblock Diffusion motion: Generate text-guided 3d human motion by diffusion
  model.
\newblock {\em arXiv preprint arXiv:2210.12315}, 2022.

\bibitem{rombach2022high}
Robin Rombach, Andreas Blattmann, Dominik Lorenz, Patrick Esser, and Bj\"orn
  Ommer.
\newblock High-resolution image synthesis with latent diffusion models.
\newblock In {\em Proceedings of the IEEE/CVF Conference on Computer Vision and
  Pattern Recognition}, 2022.

\bibitem{saharia2022photorealistic}
Chitwan Saharia, William Chan, Saurabh Saxena, Lala Li, Jay Whang, Emily
  Denton, Seyed Kamyar~Seyed Ghasemipour, Burcu Karagol~Ayan, S.~Sara Mahdavi,
  Rapha~Gontijo Lopes, Tim Salimans, Jonathan Ho, David~J. Fleet, and Mohammad
  Norouzi.
\newblock Photorealistic text-to-image diffusion models with deep language
  understanding.
\newblock {\em arXiv preprint arXiv:2205.11487}, 2022.

\bibitem{saremi2019neural}
Saeed Saremi and Aapo Hyv{\"a}rinen.
\newblock Neural empirical bayes.
\newblock {\em Journal of machine learning research: JMLR}, 20(181):1--23,
  2019.

\bibitem{schulman2015high}
John Schulman, Philipp Moritz, Sergey Levine, Michael Jordan, and Pieter
  Abbeel.
\newblock High-dimensional continuous control using generalized advantage
  estimation.
\newblock {\em arXiv preprint arXiv:1506.02438}, 2015.

\bibitem{schulman2017proximal}
John Schulman, Filip Wolski, Prafulla Dhariwal, Alec Radford, and Oleg Klimov.
\newblock Proximal policy optimization algorithms.
\newblock {\em arXiv preprint arXiv:1707.06347}, 2017.

\bibitem{shimada2021neural}
Soshi Shimada, Vladislav Golyanik, Weipeng Xu, Patrick P{\'e}rez, and Christian
  Theobalt.
\newblock Neural monocular 3d human motion capture with physical awareness.
\newblock {\em ACM Transactions on Graphics (ToG)}, 40(4):1--15, 2021.

\bibitem{shimada2020physcap}
Soshi Shimada, Vladislav Golyanik, Weipeng Xu, and Christian Theobalt.
\newblock Physcap: Physically plausible monocular 3d motion capture in real
  time.
\newblock {\em ACM Transactions on Graphics (TOG)}, 39(6), dec 2020.

\bibitem{sinha2021d2c}
Abhishek Sinha, Jiaming Song, Chenlin Meng, and Stefano Ermon.
\newblock {D2C}: {D}iffusion-decoding models for few-shot conditional
  generation.
\newblock In {\em Advances in Neural Information Processing Systems}, 2021.

\bibitem{sohl-dickstein2015deep}
Jascha Sohl-Dickstein, Eric Weiss, Niru Maheswaranathan, and Surya Ganguli.
\newblock Deep unsupervised learning using nonequilibrium thermodynamics.
\newblock In {\em International Conference on Machine Learning}, 2015.

\bibitem{song2020denoising}
Jiaming Song, Chenlin Meng, and Stefano Ermon.
\newblock Denoising diffusion implicit models.
\newblock In {\em International Conference on Learning Representations}, 2021.

\bibitem{song2020score}
Yang Song, Jascha Sohl-Dickstein, Diederik~P. Kingma, Abhishek Kumar, Stefano
  Ermon, and Ben Poole.
\newblock Score-based generative modeling through stochastic differential
  equations.
\newblock In {\em International Conference on Learning Representations}, 2021.

\bibitem{stein1981estimation}
Charles~M Stein.
\newblock Estimation of the mean of a multivariate normal distribution.
\newblock {\em The annals of Statistics}, pages 1135--1151, 1981.

\bibitem{tevet2022human}
Guy Tevet, Sigal Raab, Brian Gordon, Yonatan Shafir, Daniel Cohen-Or, and
  Amit~H Bermano.
\newblock Human motion diffusion model.
\newblock {\em arXiv preprint arXiv:2209.14916}, 2022.

\bibitem{vahdat2020nvae}
Arash Vahdat and Jan Kautz.
\newblock Nvae: A deep hierarchical variational autoencoder.
\newblock {\em Advances in Neural Information Processing Systems},
  33:19667--19679, 2020.

\bibitem{vahdat2021score}
Arash Vahdat, Karsten Kreis, and Jan Kautz.
\newblock Score-based generative modeling in latent space.
\newblock In {\em Advances in Neural Information Processing Systems}, 2021.

\bibitem{Vincent11}
Pascal Vincent.
\newblock A connection between score matching and denoising autoencoders.
\newblock {\em Neural Computation}, 23(7):1661--1674, 2011.

\bibitem{wang2019imitation}
Borui Wang, Ehsan Adeli, Hsu-kuang Chiu, De-An Huang, and Juan~Carlos Niebles.
\newblock Imitation learning for human pose prediction.
\newblock In {\em Proceedings of the IEEE International Conference on Computer
  Vision}, pages 7124--7133, 2019.

\bibitem{won2020scalable}
Jungdam Won, Deepak Gopinath, and Jessica Hodgins.
\newblock A scalable approach to control diverse behaviors for physically
  simulated characters.
\newblock {\em ACM Transactions on Graphics (TOG)}, 39(4):33--1, 2020.

\bibitem{yan2018mt}
Xinchen Yan, Akash Rastogi, Ruben Villegas, Kalyan Sunkavalli, Eli Shechtman,
  Sunil Hadap, Ersin Yumer, and Honglak Lee.
\newblock Mt-vae: Learning motion transformations to generate multimodal human
  dynamics.
\newblock In {\em Proceedings of the European Conference on Computer Vision
  (ECCV)}, pages 265--281, 2018.

\bibitem{yi2022physical}
Xinyu Yi, Yuxiao Zhou, Marc Habermann, Soshi Shimada, Vladislav Golyanik,
  Christian Theobalt, and Feng Xu.
\newblock Physical inertial poser (pip): Physics-aware real-time human motion
  tracking from sparse inertial sensors.
\newblock In {\em Proceedings of the IEEE/CVF Conference on Computer Vision and
  Pattern Recognition}, pages 13167--13178, 2022.

\bibitem{yuan20183d}
Ye Yuan and Kris Kitani.
\newblock 3d ego-pose estimation via imitation learning.
\newblock In {\em Proceedings of the European Conference on Computer Vision
  (ECCV)}, pages 735--750, 2018.

\bibitem{yuan2019ego}
Ye Yuan and Kris Kitani.
\newblock Ego-pose estimation and forecasting as real-time pd control.
\newblock In {\em Proceedings of the IEEE International Conference on Computer
  Vision}, pages 10082--10092, 2019.

\bibitem{yuan2020dlow}
Ye Yuan and Kris Kitani.
\newblock Dlow: Diversifying latent flows for diverse human motion prediction.
\newblock In {\em Proceedings of the European Conference on Computer Vision
  (ECCV)}, pages 346--364. Springer, 2020.

\bibitem{yuan2020residual}
Ye Yuan and Kris Kitani.
\newblock Residual force control for agile human behavior imitation and
  extended motion synthesis.
\newblock In {\em Advances in Neural Information Processing Systems}, 2020.

\bibitem{yuan2021simpoe}
Ye Yuan, Shih-En Wei, Tomas Simon, Kris Kitani, and Jason Saragih.
\newblock Simpoe: Simulated character control for 3d human pose estimation.
\newblock In {\em Proceedings of the IEEE/CVF Conference on Computer Vision and
  Pattern Recognition (CVPR)}, 2021.

\bibitem{zell2017joint}
Petrissa Zell, Bastian Wandt, and Bodo Rosenhahn.
\newblock Joint 3d human motion capture and physical analysis from monocular
  videos.
\newblock In {\em Proceedings of the IEEE Conference on Computer Vision and
  Pattern Recognition Workshops}, pages 17--26, 2017.

\bibitem{zeng2022lion}
Xiaohui Zeng, Arash Vahdat, Francis Williams, Zan Gojcic, Or Litany, Sanja
  Fidler, and Karsten Kreis.
\newblock {LION}: {L}atent point diffusion models for {3D} shape generation.
\newblock In {\em Advances in Neural Information Processing Systems}, 2022.

\bibitem{zhang2022motiondiffuse}
Mingyuan Zhang, Zhongang Cai, Liang Pan, Fangzhou Hong, Xinying Guo, Lei Yang,
  and Ziwei Liu.
\newblock Motiondiffuse: Text-driven human motion generation with diffusion
  model.
\newblock {\em arXiv preprint arXiv:2208.15001}, 2022.

\bibitem{zhang2022fast}
Qinsheng Zhang and Yongxin Chen.
\newblock Fast sampling of diffusion models with exponential integrator.
\newblock {\em arXiv preprint arXiv:2204.13902}, 2022.

\bibitem{zhang2022gddim}
Qinsheng Zhang, Molei Tao, and Yongxin Chen.
\newblock {gDDIM}: {G}eneralized denoising diffusion implicit models.
\newblock {\em arXiv preprint arXiv:2206.05564}, 2022.

\bibitem{zhou2021shape}
Linqi Zhou, Yilun Du, and Jiajun Wu.
\newblock {3D} shape generation and completion through point-voxel diffusion.
\newblock In {\em Proceedings of the IEEE/CVF International Conference on
  Computer Vision}, 2021.

\bibitem{zhuang2022music2dance}
Wenlin Zhuang, Congyi Wang, Jinxiang Chai, Yangang Wang, Ming Shao, and Siyu
  Xia.
\newblock Music2dance: Dancenet for music-driven dance generation.
\newblock {\em ACM Transactions on Multimedia Computing, Communications, and
  Applications (TOMM)}, 18(2):1--21, 2022.

\end{thebibliography}
